\documentclass[conference]{IEEEtran}

\usepackage{blindtext}
\usepackage{eso-pic}
\IEEEoverridecommandlockouts
\usepackage{cite}
\usepackage{amsmath,amssymb,amsfonts}
\usepackage{algorithmic}
\usepackage{graphicx}
\usepackage{textcomp}
\usepackage{xcolor}

\usepackage[export]{adjustbox}
\usepackage[utf8]{inputenc} 
\usepackage[T1]{fontenc}    
\usepackage{hyperref}       
\usepackage{url}            
\usepackage{booktabs}       
\newcommand{\tabitem}{~~\llap{\textbullet}~~}
\usepackage{makecell}
\usepackage{amsfonts}       
\usepackage{nicefrac}       
\usepackage{microtype}      
\usepackage{xcolor}         
\usepackage{amsmath,amssymb,multicol,latexsym}
\usepackage{pgfplots,pgfplotstable}
\pgfplotsset{compat=1.16}
\usepackage{acro}
\usepackage{tikz}
\usepackage{comment}
\usetikzlibrary{shapes,
 	automata,
 	arrows.meta,arrows,
 	chains,
 	matrix,
 	backgrounds,
 	fit,
 	patterns,
 	decorations.markings,
 	svg.path,
 	shapes.multipart,
 	shapes.symbols,
 	external,
 	shadows,
 	positioning,
 	calc,
 	decorations,
 	scopes,
 	patterns,
 	patterns.meta,
 	bending}
 \usetikzlibrary{external}

\usepackage{hyperref}
\usepackage{cleveref}
\usepackage{flushend}

\DeclareMathAlphabet{\mathcal}{OMS}{cmsy}{m}{n}

\usepackage{url}
\usepackage{graphicx}
\usepackage{svg}
\usepackage{numprint,fullpage}
\usepackage{booktabs}
\usepackage{multirow}

\usepackage{adjustbox}

\Crefname{equation}{Eq.}{Eqs.}
\Crefname{figure}{Fig.}{Figs.}
\Crefname{tabular}{Tab.}{Tabs.}
\usepackage{tabu}
\usepackage{anysize} 
\graphicspath{{figures/}}
\def\BibTeX{{\rm B\kern-.05em{\sc i\kern-.025em b}\kern-.08em
    T\kern-.1667em\lower.7ex\hbox{E}\kern-.125emX}}
\usepackage{url}

\DeclareAcronym{ODD}{
    short = ODD,
    long = operational design domain
}

\DeclareAcronym{PG}{
    short = PG,
    long = procedural generation
}

\DeclareAcronym{VAE}{
    short = VAE,
    long = variational autoencoder,
}

\DeclareAcronym{RL}{
    short = RL,
    long = reinforcement learning
}

\DeclareAcronym{QL}{
    short = QL, 
    long = Q-Learning
}

\DeclareAcronym{DQN}{
    short = DQN, 
    long = Deep Q-Network
}

\DeclareAcronym{A2C}{
    short = A2C, 
    long = Advantage Actor Critic
}

\DeclareAcronym{TRPO}{
    short = TRPO, 
    long = Trust Region Policy Optimization
}

\DeclareAcronym{ADS}{
    short = ADS, 
    long = automated driving system
}

\DeclareAcronym{DNN}{
    short = DNN, 
    long = deep neural network
}

\DeclareAcronym{SUT}{
    short = SUT, 
    long = system under test
}

\DeclareAcronym{NDD}{
    short = NDD, 
    long = naturalistic driving data
}

\begin{document}

\title{1001 Ways of Scenario Generation for Testing of Self-driving Cars: A Survey}


\author{\IEEEauthorblockN{Barbara~Schütt, Joshua~Ransiek, Thilo~Braun, Eric~Sax}
\IEEEauthorblockA{
FZI Research Center for Information Technology \\
Karlsruhe, Germany
Email: \{schuett, ransiek, braun, sax\}@fzi.de}}
\maketitle


%


\maketitle

\begin{abstract}
Scenario generation is one of the essential steps in scenario-based testing and, therefore, a significant part of the verification and validation of driver assistance functions and autonomous driving systems.
However, the term \textit{scenario generation} is used for many different methods, e.g., extraction of scenarios from naturalistic driving data or variation of scenario parameters.
This survey aims to give a systematic overview of different approaches, establish different categories of scenario acquisition and generation, and show that each group of methods has typical input and output types.
It shows that although the term is often used throughout literature, the evaluated methods use different inputs and the resulting scenarios differ in abstraction level and from a systematical point of view.
Additionally, recent research and literature examples are given to underline this categorization.
\end{abstract}



%
\IEEEpeerreviewmaketitle

\section{Introduction}
\label{sec:introduction}
The importance of scenario-based testing for highly automated driving made a noticable jump in 2022. 
The implementation regulation \cite{EUtypeapprovannex} regarding EU Regulation No. 2019/2144 \cite{EUtypeapprov} came into effect on 5 August 2022, and is now mandatory to be followed in all EU member states. 
It lists a number of methods for the overall compliance assessment of an automated driving system. 
The implementation regulation states that a minimum set of traffic scenarios shall be used in case these scenarios are relevant to the \ac{ADS}'s \ac{ODD}. 
For some functions, e.g., lane keep assistance, it even defines the minimum test scenarios. 
However, in general, the scenarios to be used in the development and testing phases of driving functions and assistance systems must be found, created, or generated.

A significant amount of research has been dedicated to finding and generating new scenarios for testing. 
Many of these efforts fall under the umbrella term \textit{scenario generation}, which encompasses a wide range of different methods. 
However, the concepts underlying their generation and the types of scenarios produced can vary significantly, e.g., one approach may involve creating new trajectories \cite{abeysirigoonawardena2019generating}, while another generates scenarios with new sequences of atomic scenarios and actor behavior \cite{goss2021generation}. 
In order to structure and better understand this rapidly growing field, this work proposes a categorization for different generation methods, using the term \textit{scenario acquisition} as a more general expression for obtaining new scenarios. 
The term \textit{scenario generation} is used for a special category of scenario acquisition, further explained in \cref{sec_scenario_generation}. 
This survey aims to provide a systematic overview of different approaches, establish different categories of scenario acquisition and generation, and highlight that each group of methods has typical input and output types.
It demonstrates that although the term is often used throughout the literature, the evaluated methods use different inputs and the resulting scenarios differ in abstraction level and from a systematic point of view.

\begin{figure}[t!]
    \centering
       \includegraphics[width=0.92\linewidth]{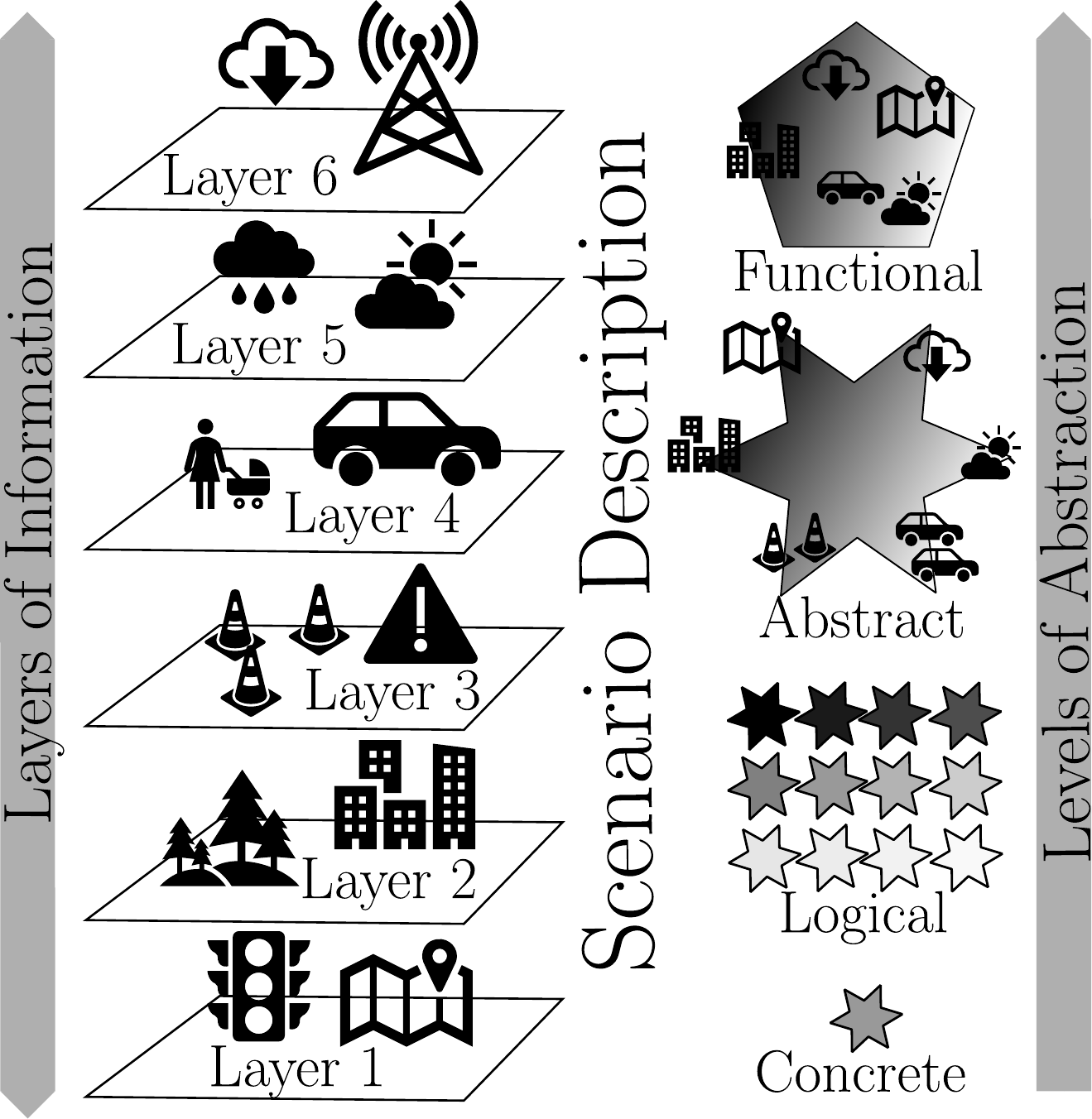}
\caption{Summary of two different scenario description concepts: scenario information layers \cite{bagschik2018ontology, scholtes20216} and scenario abstraction levels \cite{menzel2018scenarios, neurohr2021criticality}.}
\label{fig:sota}
\vspace{-3ex}
\end{figure}

The main contributions of this work are:
\begin{itemize}
    \item A scenario-driven categorization for different types of scenario acquisition (see \Cref{fig:scenario_elicitation}),
    \item a clarification of the relationship between different scenario abstraction levels and generation methods,
    \item a summary of how the proposed categories can be combined (\Cref{fig:scenario_conversion}),
    \item examples for each category from recent research and literature.
\end{itemize}

\Cref{sec:sota} gives a short introduction into important concepts, i.e., scenario abstraction levels and the 6-layer model.
\Cref{sec:taxonomy} describes the proposed categories with examples from recent literature and how these categories can be combined to enhance the results, followed by a discussion in \Cref{sec:discussion} and \Cref{sec:conclusion} which concludes this work.

\section{Background}
\label{sec:sota}
The categorization proposed in this work is based on two major scenario-related concepts, the 6-layer model and the four scenario abstraction levels of scenario information description.
While the layers group information that composes a scenario, the abstraction levels specify how abstractly this information is given.
Both concepts are shown in \Cref{fig:sota}, where the left side depicts the information layers and the right side the abstraction levels.

\subsection{Scenario Description Layers}
A scenario contains different types of information and aspects defining it as shown on the left side of \Cref{fig:sota}.
In most cases, only parts of every possible aspect within a scene or scenario are relevant to the intended use case. 
Depending on the test case, parts of a scenario can be neglected. 
Not all test cases require all aspects of a scenario description, such as weather or road friction.
To categorize the different aspects of a scenario, the PEGASUS Project \cite{bagschik2018ontology} introduced a layer model for scenarios.
According to this model, a scenario's information can be divided into six layers.
Layers 1 and 2 consist of the road and traffic infrastructure, respectively.
Layer 3 holds information about temporary changes in layer 1 and 2, e.g., construction sights.
Objects, e.g., cars or animals, are part of layer 4, and environmental information, e.g., weather or light conditions, of layer 5.
The last layer, layer 6, contains digital information, e.g., available communication networks \cite{scholtes20216}.

\subsection{Scenario Abstraction Levels}
\label{sec:scenario_abstraction}
According to Menzel \textit{et al.} \cite{menzel2018scenarios}, scenarios can be categorized into different levels of abstraction (functional, logical, and concrete scenarios) depending on their stage during the development and testing phase and get more and more refined through the development and testing process.
	
The most abstract level of scenario representation is called \textit{functional} and describes a scenario via linguistic notation using natural, non-structured language terminology. 
This level's primary goal is to create easily understandable scenarios for experts.
It describes the base road network and all actors and their maneuvers, such as a right-turning vehicle and road-crossing bike.
The next abstraction level is \textit{logical} scenarios and refines functional scenarios by adding parameters. 
These parameters can, for instance, be ranges for road width, vehicle positions and velocity, or time and weather conditions.
The parameters can either be described as parameter ranges or probability distributions.
The most detailed level is \textit{concrete} scenarios, which depict operating scenarios with concrete values for each parameter in the parameter space.	
One logical scenario can yield many concrete scenarios, depending on the number of variables, width, and step size for these ranges.

Neurohr \textit{et al.} \cite{neurohr2021criticality} introduce an additional abstraction level between the functional and logical levels - the \textit{abstract} scenario.
This level consists of formalized and machine-readable scenario descriptions. 
It is closely related to an ontology to describe relations within a scenario efficiently.
A conceptual overview of all four levels of  abstraction is given in \Cref{fig:sota} on the right side.

\subsection{Research Field of Scenario Generation}
There already exist a number of survey articles and reviews on scenario generation.
One approach for structuring this field is an algorithm-based method \cite{ding2022survey}, where different scenario generation methods are grouped regarding the utilized type of generation, i.e., data-driven, adversarial, and knowledge-based.
An alternative approach is to summarize according to the underlying data-driven model \cite{cai2022survey} or goal-oriented \cite{sun2021scenario}.
An in-detail survey for finding critical scenarios is given by Zhang \textit{et al.} \cite{zhang2021finding}.
Their work includes three main aspects: defining why scenarios need to be found, developing methods to find these scenarios, and assessing the established scenarios. 
However, their taxonomy does not differentiate between different abstract elicitation concepts, and thus can be seen as an extension of their solution level.
\section{Categories of Scenario Acquisition}
\label{sec:taxonomy}

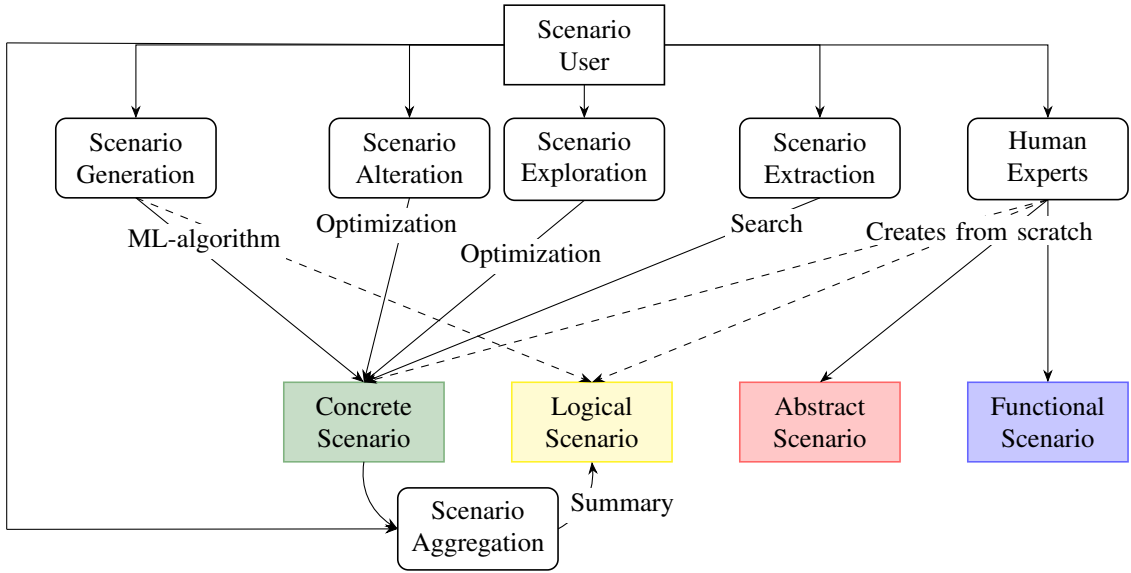
\begin{figure*}[ht!]
	\centering
	\begin{tikzpicture}[sideArrow/.style={signal, font=\small, fill=#1!20, signal pointer angle=100, align=center},]
\tikzstyle{box}=[draw, minimum height=30pt, minimum width=60pt, semithick, inner sep=5pt, outer sep=0pt, anchor=north];
\tikzstyle{process}=[draw,rounded corners, minimum height=30pt, minimum width=60pt, semithick, inner sep=5pt, outer sep=0pt, anchor=north];
\tikzstyle{textbox}=[minimum width=10pt, semithick, inner sep=2pt, outer sep=0pt, anchor=north, fill=white];
\node[box, align=center, anchor=north] at (-0.1,2.5) (Start){Scenario \\ User};


\node[process, align=center, anchor=north] at (-6,1.0) (ScenGen){Scenario\\Generation};
\draw[to path={-| (\tikztotarget)}](Start.west) edge[-{Stealth[fill=black]}] (ScenGen);

\node[process, align=center, anchor=north] at (-2.4,1.0) (ScenAlt){Scenario\\Alteration};
\draw[to path={-| (\tikztotarget)}] (Start.west) edge[-{Stealth[fill=black]}] (ScenAlt);

\node[process, align=center, anchor=north] at (-0.1,1.0) (ScenExpl){Scenario\\Exploration};
\draw[](Start.south) edge[-{Stealth[fill=black]}] (ScenExpl.north);

\node[process, align=center, anchor=north] at (3,1.0) (ScenExtr){Scenario\\Extraction};
\draw[to path={-| (\tikztotarget)}](Start) edge[-{Stealth[fill=black]}] (ScenExtr);

\node[process, align=center, anchor=north] at (6,1.0) (Experts){Human\\Experts};
\draw[to path={-| (\tikztotarget)}](Start) edge[-{Stealth[fill=black]}] (Experts);

\node (ctrl1) [anchor=center]at (-7.7,2.0) {}; 
\node (ctrl2) [anchor=center]at (-7.7,-4.45) {}; 
\node[process, align=center, anchor=north]  at (-1.5,-3.9) (sum) {Scenario\\Aggregation};
\draw[to path={-- (\tikztotarget)}](Start.west) edge[-] (ctrl1.center);
\draw[to path={-- (\tikztotarget)}](ctrl1.center) edge[-] (ctrl2.center);
\draw[to path={-- (\tikztotarget)}](ctrl2.center) edge[-{Stealth[fill=black]}] (sum.west);

\node[box, align=center, draw=black!60!green!50,fill=black!60!green!22] at (-3,-2.5) (concrete){Concrete\\Scenario};
\node[box, align=center, draw=yellow!90, fill=yellow!22] at (0,-2.5) (logical){Logical\\Scenario};
\node[box, align=center, draw=red!60, fill=red!22] at (3,-2.5) (abstract){Abstract\\Scenario};
\node[box, align=center, draw=blue!60, fill=blue!22] at (6,-2.5) (functional){Functional\\Scenario};

\draw[] (Experts.south) edge[-{Stealth[fill=black]}] (functional.north);
\draw[] (Experts.south) edge[-{Stealth[fill=black]}] (abstract.north);
\draw[dashed] (Experts.south) edge[-{Stealth[fill=black]}] (logical.north);
\draw[dashed] (Experts.south) edge[-{Stealth[fill=black]}] (concrete.north);
\node[textbox, align=center, anchor=north] at (5.1,-0.3)(Eprt2Text){Creates from scratch};

\draw[] (ScenAlt.south) edge[-{Stealth[fill=black]}] (concrete.north);
\node[textbox, align=center, anchor=north] at (-2.7,-0.2)(Alt22Text){Optimization};
\draw[] (ScenExpl.south) edge[-{Stealth[fill=black]}] (concrete.north);
\node[textbox, align=center, anchor=north] at (-0.8,-0.6)(Exol2Text){Optimization};
\draw[] (ScenExtr.south) edge[-{Stealth[fill=black]}] (concrete.north);
\node[textbox, align=center, anchor=north] at (2.3,-0.2)(Extr2Text){Search};
\draw[dashed] (ScenGen.south) edge[-{Stealth[fill=black]}] (logical.north);
\draw[] (ScenGen.south) edge[-{Stealth[fill=black]}] (concrete.north);
\node[textbox, align=center, anchor=north] at (-5.1,-0.4)(Gen2Text){ML-algorithm};
\draw[] (sum.east) edge[-{Stealth[fill=black]}, bend right=35](logical.south);
\draw[] (concrete.south) edge[-{Stealth[fill=black]}, bend right=35](sum.west);
\node[textbox, align=center, anchor=north] at (0.4,-3.9)(Agg2Text){Summary};

\end{tikzpicture}
	\caption{Context diagram of the six proposed main scenario acquisition categories and their major results of scenario types. Solid arrows show the primary occurring output scenario type, dashed arrows show less common output types.}
	\label{fig:scenario_elicitation}
 \vspace{-3ex}
\end{figure*}

Finding new scenarios is an elementary step for defining test cases to assess an \ac{ADS}'s safety since scenarios are the foundation of scenario-based testing.
Therefore, numerous concepts, methods, and algorithms are proposed throughout the literature.
Many scenario generation algorithms modify or generate new scenarios modifying information on layer 4.

In order to obtain a classification and definition of these methods, we present the following taxonomy.
It is intended to show significant conceptual differences between individual approaches and to group them accordingly.
In addition, our analysis is illustrated with examples from the current literature. 
Scenario acquisition can be grouped into six major categories, and \Cref{fig:scenario_elicitation} shows a schematic overview in a context diagram of all different categories.
Some groups can be sub-categorized further, but since it does not make any differences in the input or output scenario abstraction, these sub-categories are not shown.
The first step is to define \textbf{what} needs to be tested since the goal defines the needed scenarios.
In \Cref{fig:scenario_elicitation}, the \textbf{what} comes from the scenario user, who can be a developer or tester.
From that step, different ways of acquiring new scenarios are possible (white boxes).  
Solid arrows between the categories and scenario types show the primary occurring output scenario type.
Dashed arrows show that these outputs are also generated but less common. 
It does not mean that a scenario type is not possible as output in case there is no arrow.
However, no or few examples could be found in recent literature.
We propose the following six categories:

\begin{itemize}
    \item[1] Scenario Generation: New scenarios or sub-sequences of existing scenarios are (automatically) generated for fulfilling a given set of criteria. The primary characteristic of this category is the creation of new scenarios with novel content that do not necessarily belong to the same logical scenario.
    \item[2] Scenario Alteration: a parameter value is altered to find other related scenarios. The initial scenario does not necessarily start with a predefined parameter range but with an initial value that is changed during the search.
    \item[3] Scenario Exploration: The parameter ranges or distributions of a given logical scenario are explored for fulfilling a given set of criteria, e.g., are explored by optimization algorithms. The new values are defined by a step size within this range or drawn from a given distribution for a parameter.
    \item[4] Scenario Extraction: Recorded data is searched for specific features to extract scenarios with the wanted criteria, e.g., by machine learning algorithms or rule-based.
    \item[5] Experts: Human experts define a scenario, e.g., by drawing the scenario sequence as images or modeling concrete scenarios directly in simulation software.
    \item[6] Scenario Aggregation: Concrete scenarios found by any acquisition method can be grouped into logical or related scenarios, i.e., scenarios that share features and content for clustering or grouping.
\end{itemize}

Another aspect being considered is the respective relationship between input and output scenarios, building blocks, or data for each category.
\Cref{fig:scenario_conversion} shows an overview of possible combinations of methods and how they can increase the set of already existing scenarios and usual multiplicities for input and output types for each acquisition method.
Additionally, the found scenarios can be used for testing purposes and other applications, e.g., training automated driving functions with reinforcement learning \cite{abeysirigoonawardena2019generating}.
The found methods in this work are not intended to be exhaustive.

\subsection{Scenario Generation}
\label{sec_scenario_generation}
The first concept, scenario generation, describes creating new scenarios, e.g., a new sequence of actor states.
Scenario generation includes changing the behavior and number of actors within a scenario, or changing the sequence or tasks actors have to fulfill. 
More generally, more than predefined scenario parameters are changed. 
Methods within this category can find concrete scenarios that do not necessarily belong to the same logical scenario. 
Scenario generation methods are mainly divided into two categories. 
The first category generates scenarios from scratch, where changes on all six scenario layers are possible. 
The second category, data-driven, works with existing actor behaviors and road networks gathered from real-world data. 
These methods mostly consider modifications on layers 2 and above. 
Scenario generation from scratch can start by creating a road network (layer 1) and subsequently adding objects in the following layers, most commonly layer 4. 

\subsubsection{Scenario Generation from Scratch}
Generating scenarios from scratch usually starts with a set of scenario building blocks defined by the framework or task at hand.
These blocks are combined and changed with a generation algorithm to get new (concrete) scenarios.
Additionally to layer 4, it often uses map or road generation to add changes on layer 1 \cite{li2022metadrive, rietsch2022driver, paranjape2020modular}, which is a major difference to most other scenario acquisition types.
Used algorithms are, for example, \ac{PG} in the MetaDrive \cite{li2022metadrive} frame work, \ac{RL} by DriverDojo \cite{rietsch2022driver}, or a procedural approach \cite{paranjape2020modular}.
A specialty of MetaDrive \cite{li2022metadrive}, that should be mentioned is, that it has modules to make changes on layer 1-4.
Other approaches are keyword-based\cite{menzel2019functional} or ontology-based \cite{bagschik2018ontology}.
\subsubsection{Data-driven Scenario Generation}
Using real-world data for scenario generation usually starts with recorded concrete or logical scenarios. 
After processing, existing objects can be modified, new trajectories can be assigned, or additional static and dynamic obstacles can be added. 

Data-driven scenario generation approaches also occur mainly on layer 4 and use different generative algorithms, e.g., autoregressive generative model with an encoder-decoder architecture \cite{feng2022trafficgen}, from real-data extracted atomic block for recombination \cite{goss2021generation}, or maneuver and sequence probability \cite{ghodsi2021generating}.



Both \cite{suo2021trafficsim} and \cite{tan2021scenegen} address the problem of generating realistic traffic scenes and solely operate on layer 4. 
The former uses a  convolutional LSTM that inserts objects of various classes (vehicles, pedestrians, bicyclists) into the scene and synthesizes their sizes, orientations, and velocities.
The latter uses motion forecasting and an implicit latent variable model to generate multiple scene-consistent samples of non-ego vehicle trajectories in parallel. 

\begin{figure*}[th!]
	\centering
	\begin{tikzpicture}[sideArrow/.style={signal, font=\small, fill=#1!20, signal pointer angle=100, align=center},]
\tikzstyle{box}=[draw,rounded corners, minimum height=30pt, minimum width=70pt, semithick, inner sep=5pt, outer sep=0pt, anchor=north];
\tikzstyle{textbox}=[minimum width=10pt, semithick, inner sep=2pt, outer sep=0pt, anchor=north, fill=white];

\node[box, align=center, draw=black!60!green!50,fill=black!60!green!22] at (0,0) (concrete){Concrete Scenario};
\node[box, align=center, draw=yellow!90, fill=yellow!22] at (5.5,0) (logical){Logical Scenario};
\node[box, align=center, draw=purple!60] at (-5.5,1) (Naturalistic){Recorded Data, \\ Reports, etc.};
\node[box, align=center, draw=violet!60] at (-5.5,-1) (Building){Scenario Building \\ Blocks \& Processes};

\draw[] (Naturalistic.north) edge[-{Stealth[fill=black]}, bend left=90](concrete.north);
\node[textbox, align=center] at (-5.0,1.4)(extract){1..n};
\node[textbox, align=center] at (-0.3,0.35)(extract){n};

\draw[] (logical.north) edge[-{Stealth[fill=black]}, bend right=90](concrete.30);
\node[textbox, align=center] at (5.1,0.4)(extract){1};
\node[textbox, align=center] at (1.2,0.35)(extract){n};

\draw[] (concrete.330) edge[-{Stealth[fill=black]}, bend right=90](logical.south);
\node[textbox, align=center] at (5.1,-1.1)(extract){1};
\node[textbox, align=center] at (1.2,-1.1)(extract){n};

\draw[] (Building.south) edge[-{Stealth[fill=black]}, bend right=90](concrete.south);
\node[textbox, align=center] at (-0.25,-1.1)(extract){n};
\node[textbox, align=center] at (-5.0,-2.1)(extract){1..n};

\draw[-{Stealth[fill=black]}] (concrete.-180) arc (5:334:6mm);
\node[textbox, align=center] at (-1.5,-1.15)(extract){1..n};
\node[textbox, align=center] at (-1.6,0.4)(extract){1};

\node[textbox, align=center] at (-3,2.75)(extract){Scenario\\Extraction\\\ref{sec_scenario_extraction}};
\node[textbox, align=center] at (3.2,2.0)(explore){Scenario\\Exploration\\\ref{sec_scenario_exploartion}};
\node[textbox, align=center] at (3.2,-2.0)(sum){Scenario\\Aggregation\\\ref{sec_aggregation}};
\node[textbox, align=center] at (-2.8,0.7)(alter){Scenario\\Alteration\\\ref{sec_alteration}};
\node[textbox, align=center] at (-3.1,-2.7)(generate){Scenario\\Generation\\\ref{sec_scenario_generation}};
\end{tikzpicture}
	\caption{The input and output relationships of the proposed categories, their relations, and multiplicities.}
	\label{fig:scenario_conversion}
 \vspace{-3ex}
\end{figure*}

\subsection{Scenario Alteration and Exploration}

Scenario alteration and exploration are based on a similar concept: new concret scenarios are derived from existing scenarios. 
However, both have different points of view and initial settings.
As shown in \Cref{fig:scenario_conversion}, scenario alteration usually starts with a concrete scenario where a defined value is altered until a predefined goal is achieved, resulting in one or more new concrete scenarios.
In contrast to alteration, scenario exploration starts with a logical scenario where drawing values from parameter ranges and distributions are optimized towards a goal, e.g., criticality. 
The result of scenario exploration is a set of concrete scenarios derived from one logical scenario.
Thus, the main difference between both methods is the starting point, i.e., the definition and abstraction level of the start scenario.
The results usually are concrete scenarios for both categories. 
Unlike scenario exploration, however, the resulting concrete scenarios from scenario alteration do not necessarily have to be within one logical scenario \cite{wang2021advsim}.
Therefore, in \Cref{fig:scenario_elicitation}, both categories are grouped.
The following section shows a selection of scenario alteration and exploration examples.

\subsubsection{Scenario Alteration}
\label{sec_alteration}
Scenario alteration describes methods where parameters and scenario parts of different scenario description layers of one concrete scenario are changed.
The most common changes appear on layer 4 of the 6-layer model for scenarios, which contains dynamic objects and their movement. 
These variations mainly include trajectory alterations of different kinds of actors, e.g., pedestrians or adversary vehicles.
Different algorithms are utilized (algorithm-based), e.g., \ac{RL} methods like \ac{DQN} \cite{karunakaran2020efficient}, Bayesian optimization (BO) \cite{abeysirigoonawardena2019generating, wang2021advsim, gangopadhyay2019identification}, reachable sets (RS) \cite{althoff2018automatic}, RRT$\ast$ path planning \cite{zhang2020multi}, trajectory and velocity planner \cite{zhang2021trajectory}, evolutionary algorithm \cite{klischat2019generating}, or genetic algorithms (GA) \cite{wang2021advsim}. 
\cite{rempe2022generating} uses a graph-based conditional variational autoencoder. 
Therefore, the trajectory alteration is formulated as an optimization in the latent space. 
Adaptive stress testing \cite{koren2018adaptive} uses Trust Region Policy Optimization (TRPO) to alter pedestrian trajectories and inject noise on position and velocity measurement to find likely failure scenarios.

Even though these are the dominating generation methods in the literature, others shall also be mentioned:
A maneuver-based (information-based) variation method \cite{nonnengart2019crisgen}, a road geometry approach based on A$\ast$ and the Ant Colony Optimization \cite{ponn2020automatic}, or a road network generating framework \cite{muktadir6procedural}.
Road generation combined with platforms like Foretellix's V-Suites \cite{foretellix}, where a scenario can be executed at different suitable places on a map, might be used in future work.

\subsubsection{Scenario Exploration}
\label{sec_scenario_exploartion}
Another sub-category is scenario exploration. 
Sometimes it is not easy to distinguish between scenario exploration and alteration without nomenclature as logical and concrete scenarios used by the authors.
Therefore, we define \textit{scenario exploration} as an exploration of a predefined parameter range or distribution defining behavior within a scenario.
Only if this constraint for parameters is fulfilled a method falls into the category of scenario exploration.
Otherwise, it shall be labeled as scenario alteration.

The main goal in this category is to explore the given parameter range or distribution to find concrete scenarios that fulfill a given set of criteria, e.g., scenario criticality metrics \cite{westhofen2022criticality, schutt2022fingerprint}.
In this case, a derived set of concrete scenarios (output) is the wanted result of a given logical scenario (input).
A typical approach is to use optimization algorithms: evolutionary algorithms \cite{bussler2020application, abdessalem2018testing}, various \ac{RL} algorithms \cite{ding2020learning}, e.g., \ac{QL} \cite{baumann2021automatic}, \ac{A2C} \cite{kuutti2020training}, random parameter generation and evaluation of severity and newness \cite{thal2022generic}, or criticality as an optimization problem with Bayesian optimization and Gaussian processes \cite{schutt_application_2022}. 
In general, these approaches are combined with different criticality metrics, e.g., distance or time-to-collision, which are optimized so that they shall be as critical as possible within a parameter range or distribution or the resulting scenarios are categorized regarding their criticality and severity \cite{wang2020behavioral}.

For a complete picture, Mori \textit{et al.} \cite{mori2022inadequacy} show in their experiments that a discrete set of predefined parameters within a given range can yield problems.
In their work, they suggest that for a standard \ac{DNN} perception algorithm, found outliers and fluctuations lead to unpredictable behavior of the \ac{SUT} and high error rate if the grid results are interpolated to the continuous space.
As expected, this error rate grows with a higher grid step size but does not decrease after a certain small step size threshold.
According to them, the main issue is that these outliers and fluctuations might be overlooked during scenario-based testing.
In their opinion, distribution-based sampling or optimization over parameter ranges might solve this problem.

Additionally, optimization-based methods cannot be sure to find the best possible, e.g., the most critical, scenario existing in parameter space.
Hungar \cite{hungar2020concept} delineates an analytical approach to how the combination of Lipschitz-based extrapolations and systematic, semantics-guided search for extrapolation boundaries might overcome this deficiency.

The simulation of scenarios during the optimization is not necessary in every case:
Scenario evaluation using surrogate models is an alternative to data-driven scenario derivation and evaluating scenarios after simulation execution \cite{Abdessalem2016testingNN}.
Instead of possible expensive physics-based simulations, Abdessalem \textit{et al.} \cite{Abdessalem2016testingNN} proposed a surrogate model for each fitness function, i.e., the minimal distance between ego and pedestrian, the minimal distance between pedestrian and a warning area of the ego vehicle's sensor, and minimum time-to-collision, to predict the criticality metric results without running the actual simulations.
These surrogate models can be developed and trained using a combination of multi-objective search and neural networks and are then used to train and evaluate different evolutionary algorithms.

\subsection{Scenario Extraction}
\label{sec_scenario_extraction}
The next category is extracting scenarios from recorded driving data. 
Usually, this method results in concrete scenarios as shown in \Cref{fig:scenario_conversion}.
In many cases, the given data sets are \ac{NDD} sets \cite{interactiondataset, inDdataset, highDdataset}.

\subsubsection{Extraction from Naturalistic Driving Data}
Rule-based extraction uses predefined rules to extract features from the data describing the scenario.
This can exemplary be maneuvers \cite{tenbrock2021conscend, braun2021collection, king2021capturing, hartjen2019semantic, zofka2015data, kerber2020clustering}, occupancy grids \cite{weber2022unscene, balasubramanian2021traffic}, or features from the static environment \cite{langner2019logical}, dynamic time warping combined with principal component analysis \cite{hauer2020clustering}, scenario scene graph based extraction \cite{tottel2022reliving}.
Since \ac{NDD}s contain a concrete scenario usually more than once, it is a frequent practice to combine this method with a subsequent scenario aggregation to create logical scenarios \ref{sec:combination of methods}.
Furthermore, recorded data can be used to find unknown scenarios. 
Thereby, unsupervised machine learning methods are applied to find anomalies in traffic scenarios.
Langner \textit{et al.} \cite{langner2018estimating} propose autoencoders to identify previously unknown scenarios. Based on an initial set, the algorithm detects novel scenarios automatically and adds them to the known set of scenarios.

\subsubsection{Crash Data}
Another data source for scenario extraction are catalogs of accident data. 
In contrast to naturalistic driving data, accident databases only contain preselected critical scenarios. These scenarios can be extracted from various sources, such as crash videos using Deep Neural Networks (DNNs) \cite{xinxin2020csg}, natural language processing of police reports \cite{gambi2019generating}, or by analyzing crash reports and using logistic regression equations \cite{esenturk2021analyzing}. 
For example, Cao \textit{et al.} \cite{cao2019typical} extracted typical pre-crash scenarios between two-wheelers and passenger vehicles from a catalog of 216 accident scenarios.




\subsection{Scenario Creation by Experts}
This category is worth mentioning but is outside the scope of this work. 
Human experts can create scenarios of all abstraction levels.
In particular functional scenarios are human-readable and at an abstraction level for human understanding and less addressed to be used for machine interpretation \cite{menzel2018scenarios}.
They can be used during the concept phase of the ISO 26262 \cite{iso26262} standard, ensuring easy understanding and discussion.
The abstract level includes scenarios that can be interpreted by (simulation) software but primarily define constraints and dependencies within a scenario and not necessarily concrete values.
Examples for the functional and abstract level are graphical description languages, such as theater/movie storyboard related \cite{damm2017traffic}, directed graph \cite{schutt2020sceml}, or behavior trees \cite{Dosovitskiy17, bauer2021yase}.
Most simulation software vendors offer different scenario modeling interfaces, e.g., dSpace \cite{dSpacesim}, CarMaker \cite{IPGsim}, Foretellix \cite{foretellix}, or VDT \cite{VTDsim}. 
Some of them already support a version of the arising standard OpenSCENARIO \cite{openscenario}.

\subsection{Scenario Aggregation}
\label{sec_aggregation}
Scenario aggregation can be used to group similar concrete scenarios into one logical scenario and is often a second step after a set of concrete scenarios is acquired.
Aggregation methods can be based on different approaches: e.g., maneuver and interaction based \cite{king2021capturing}, (feature-based) clustering \cite{langner2019logical, weber2022unscene, hauer2020clustering, kerber2020clustering}, convolutional neural network combined with random forests \cite{balasubramanian2021traffic}.


\subsection{Combination of Methods}
\label{sec:combination of methods}
Several approaches include a combination of methods, e.g. when concrete scenarios are aggregated after the extraction \cite{weber2022unscene, braun2022maneuver, king2021capturing, langner2019logical, hauer2020clustering, kerber2020clustering}.
Other combinations are scenario extraction and expert scenario creation \cite{braun2021collection}, or scenario extraction and alteration \cite{zofka2015data}.
\Cref{fig:scenario_conversion} shows an overview of possible combinations of automated methods and how they can increase the set of already existing scenarios.
Scenario extraction from data sets or other recordings of traffic situations and scenario generation can lead to new concrete scenarios.
Scenario alteration leads to new but related scenarios of an initial scenario.
The so-found concrete scenarios can be aggregated into logical scenarios.
Scenario exploration can be utilized to find relevant or critical areas within a scenario space (logical scenario).

\section{Discussion}
\label{sec:discussion}
A qualified verification and validation strategy is needed for safety reasons. 
A basic approach is scenario-based testing, including the finding, derivation, and creation of scenarios for testing.
However, the main problem is that it is impossible to think of and test all imaginable (and unimaginable) scenarios.
Sometimes, traffic scenes and situations might occur in real life that no one was able to think about before, and of course, these are not part of any scenario database until their date of occurrence.
Most scenario acquisition methods, i.e., exploration, alteration, or extraction, are likely unable to find such scenarios. 
Nevertheless, these methods can make existing situations and scenarios more severe and thus worth testing.
On the other hand, scenario generation methods can generate new scenarios which are not necessarily based on actual events.
Therefore, a trade-off between both methods has to be found, and guidelines for what needs to be tested have to be given, e.g., the definition of a suitable \ac{ODD} to derive scenarios.
As shown in this survey, most approaches use actual events as a basis for their work.
However, the generation of entirely new scenarios is just starting to begin and leaves a lot of different areas and approaches for future work.

\section{Conclusion}
\label{sec:conclusion}

In this paper, we contribute to the field by giving a structured overview of six different scenario acquisition methods and showing how they are related and their potential for combined approaches.
This categorization is a scenario-driven approach, and the different categories are defined by their data and scenario input and output.
Some of them can be further divided into sub-categories, which do not influence their input and output scenario types.
Further, we described how these categories are related and can be combined to get the desired output format.
Additionally, our categorization is supported by examples from recent research and literature.

The proposed taxonomy can help to define suitable input and output for scenario generation, depending on the question of what should be tested, the intended test case for \ac{ADS}, and the available abstraction level of already existing scenarios.
Furthermore, combined methods can help to enhance or increase the scenario output quantity and quality.

\section*{Acknowledgment}
The research leading to these results is funded by the German Federal Ministry for Economic Affairs and Climate Action within the projects \textit{Verifikations- und Validierungsmethoden automatisierter Fahrzeuge im urbanen Umfeld} a project from the PEGASUS family, based on a decision by the Parliament of the Federal Republic of Germany. The authors would like to thank the consortium for the successful cooperation.



\bibliographystyle{IEEEtran}
\bibliography{lit}
%

\end{document}